# TRUST-BASED SYMBOLIC MOTION PLANNING FOR MULTI-ROBOT BOUNDING OVERWATCH

Huanfei Zheng[1], Jonathon M. Smereka[2], Dariusz Mikulski[2], Stephanie Roth[2], Yue Wang[1]

[1]Mechanical Engineering Department, Clemson University, Clemson, SC
[2]U.S. Army CCDC Ground Vehicle Systems Center, Warren, MI

**ABSTRACT**

*Multi-robot bounding overwatch requires timely coordination of robot team members. Symbolic motion planning (SMP) can provide provably correct solutions for robot motion planning with high-level temporal logic task requirements. This paper aims to develop a framework for safe and reliable SMP of multi-robot systems (MRS) to satisfy complex bounding overwatch tasks constrained by temporal logics. A decentralized SMP framework is first presented, which guarantees both correctness and parallel execution of the complex bounding overwatch tasks by the MRS. A computational trust model is then constructed by referring to the traversability and line of sight of robots in the terrain. The trust model predicts the trustworthiness of each robot team's potential behavior in executing a task plan. The most trustworthy task and motion plan is explored with a Dijkstra searching strategy to guarantee the reliability of MRS bounding overwatch. A robot simulation is implemented in ROS Gazebo to demonstrate the effectiveness of the proposed framework.*

## 1. INTRODUCTION

Bounding overwatch is a process of alternating movement of coordinated teams to move forward under potential adversaries [1]. As members in a team take an overwatch posture, other members advance to cover. In robotic bounding overwatch, teams of (semi)autonomous robotic ground vehicles are coordinated to perform such tasks autonomously while a human operator supervises the task and intervenes if necessary. The process of a two-robot bounding overwatch is illustrated in Fig. 1. Generally, there are two variants of bounding overwatch. One method is the successive bounding overwatch that part of robot team members moves, halt and wait for the remaining members to reach the current overwatch point. It is used when maximum security and ease of control are required. The other method is the alternating bounding overwatch that part of robot team members moves, halt and wait for the remaining members to pass the current overwatch point. It is used when security and more rapid advancement are required.

Distribution A: Approved for public release; distribution unlimited. OPSEC 3849. Research was partially supported by the Automotive Research Center under grant no. W56HZV-19-2-0001.

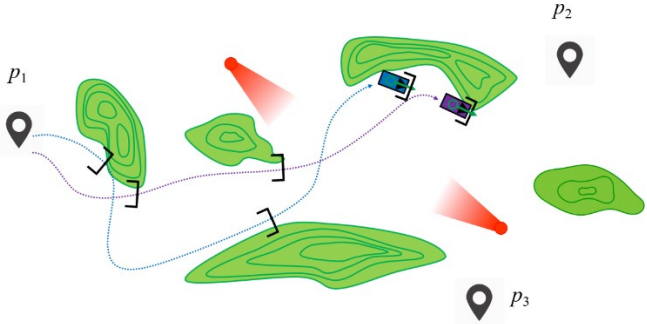

**Figure 1:** The bounding and overwatch process of a two-robot team in a terrain. Black landmarks are the desired destinations that robot team needs to visit. Red sectors are the potential static adversaries and their detection ranges.

Beyond reachability to goal destinations, temporal logic constraints (such as sequence of actions, synchronized coordination between robots) can also be critical for complex motion planning problems in bounding overwatch. However, current bounding overwatch planning is always based on a fixed sequence in visiting the overwatch points, ignoring the fact that there may exist alternative bounding overwatch plans that may satisfy human supervisors' requirements [2, 3]. It is important to determine which motion plan to choose for bounding overwatch at each step. Traditional motion planning such as A* and RRT* does not consider such problems and has difficulty in specifying complex tasks with temporal logic constraints for a robot. Symbolic motion planning (SMP) synchronizes task specifications with transition systems of robots to provide provably correct solutions to high-level goals in a discretized workspace [4]. Many centralized and decentralized frameworks have been developed for multi-robot system (MRS) SMP with model checking techniques. These works focused on reducing the computational complexity either with top-down or bottom-up strategy [7-9]. Correct task or motion planning solutions can be obtained from the model checking results of the SMP. However, limited amount of work deals with the problem of task assignment into robot.

On the other hand, there has been a lot of recent attention on trust-based decision-making for evaluating the trustworthiness of robot's behavior in human robot collaboration systems [10-15, 19, 20]. These evaluations aim to improve the robot performance or robots' physical behaviors in the human robot interaction process. However, they do not consider the selection of most trustworthy solution among multiple qualified task or motion planning solutions. Except our first attempt in [5, 6], there generally lacks discussion on the trustworthiness of task and motion plans in SMP scenarios with humans-in-the-loop, not to mention its application in multi-robot bounding overwatch.

In this paper, we propose to develop a SMP framework and computational trust models for MRS to select overwatch points in accomplishing complex missions. The complex missions are nontrivial robot motion planning tasks and associated with temporal logic requirements, which are more than travelling from one position to another. The proposed SMP framework will provide a decentralized approach for heterogeneous MRS to collaboratively achieve the complex missions with application to bounding overwatch. Compared to extant approaches, our proposed framework will improve computation efficiency and concurrency of temporal logics constrained complex mission execution. The proposed computational trust model for MRS will evaluate the trustworthiness of the decision-making of each robot team in selecting the bounding overwatch points. The trust model accommodates robot and environment uncertainties that affect human trust in MRS autonomous task performing. It takes into account a set of concrete factors (e.g., the traversability and line of sight of each robot) for the nontrivial robot planning tasks. The association of computation trust model with the SMP will generate the most trustworthy bounding overwatch task and motion plans of MRS.

## 2. FRAMWORK OF TRUST-BASED SMP FOR MRS BOUNING OVERWATCH

Assume that a set of complex motion tasks are assigned to multiple teams of heterogeneous robots in a mission domain. Each task is associated with the exploration of a key fort in the domain. The tasks are also subject to temporal logic constraints, which give the rules regarding how the tasks can be achieved in a reasonable timeline. The task specification with temporal logic constraints can be described by the following formulae in Def. 1.

**Definition 1. (LTL Specification [16])** A linear temporal logic (LTL) formula $\varphi$ is formed from atomic propositions, propositional logic operators, and temporal operators according to the grammar

$$\varphi ::= true \mid \pi \mid \neg\varphi \mid \varphi_1 \vee \varphi_2 \mid \bigcirc\varphi \mid \varphi_1 U \varphi_2,$$

where $\varphi$ is an atomic proposition, $\neg$ (negation) and $\vee$ (disjunction) are Boolean operators, and $\bigcirc$ (next) and $U$ (until) are temporal operators.

More expressive operators can be constructed from the above operators, such as, conjunction: $\varphi_1 \wedge \varphi_2 = \neg(\neg\varphi_1 \vee \neg\varphi_2)$, eventually: $\Diamond\varphi = true\ U\ \varphi$, and always: $\square\varphi = \neg\Diamond\neg\varphi$.

LTL formula can describe a high-level bounding overwatch task specification constrained by temporal logic. For example, we can use the LTL formula "$\Diamond\square\pi$" to describe the task specification for a bounding overwatch task that "finally always explore the target fort in a mission domain". Here, $\pi$ is the atomic proposition – "exploring the target fort". Similarly, regular expression (RE) formulae can also denote the temporal logic constrained tasks.

**Definition 2. (Regular Expression [16])** A regular expression (RE) over an alphabet $\Phi$ is defined as follows: (i) $\varepsilon, \phi_1, \phi_2 \in \Phi$ are called the primitive REs; (ii) the concatenation $(\phi_1 \cdot \phi_2)$, union $(\phi_1 + \phi_2)$, and Kleene star $(\phi_1^*)$ are the operations of REs; (iii) A string $\phi$ is a RE if and only if it can be derived from the primitive REs with finite numbers of application of the operations in (ii).

The LTL specification or RE formula can also be converted to an automaton format, which has the states and transitions, and is more expressive.

**Definition 3. (Deterministic Finite Automaton [16])** A deterministic finite automaton (DFA) is a tuple described as $G = (X, E, f, x_0, X_F)$, where $X$ is the sate set, $E$ is the event set, $f(x, e) = x'$ is the transition relation with $x, x' \in X, e \in E$, $x_0$ is the initial state, and $X_F$ is the final state set. A path of $G$, denoted by $\rho = e^{(0)} \cdots e^{(\tau)} \cdots e^{(T)}$, is a sequence of events satisfying $f(x_0, \rho) \in X_F$, where $e^{(\tau)} \in E$. The language generated by an automaton $G$ is $L(G) = \{\rho \in E^* \mid f(x_0, \rho) \in X_F\}$.

The language of the DFA enumerates all the task performing processes in a string format.

**Example 1.** Consider an example task specification - "first repeatedly visit fort $p_1$ at least once, then visit fort $p_2$ or $p_3$". The RE formula is $p_1 p_1^*(p_2 + p_3)$, while the LTL formula is $p_1 \wedge \bigcirc(p_1 U(p_2 \vee p_3))$. The equivalent automaton $G = (X, E, f, x_0, X_F)$ can be shown in Fig 2. The state set is $X = \{0, 2, 5\}$. The event set is $E = \{p_1, p_2, p_3\}$. The transition relation is listed as $f(0, p_1) = 2, f(2, p_1) = 2, f(2, p_2) = 5, f(2, p_3) = 5$. State 0 is the initial state $x_0$, 5 is the final state $X_F$.

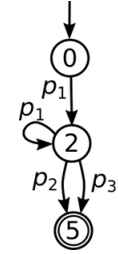

**Figure 2:** Equivalent automaton of regular expression $p_1 p_1^*(p_2 + p_3)$ and LTL $p_1 \wedge \bigcirc(p_1 U(p_2 \vee p_3))$.

**Definition 4. (Capability Markov Decision Process (MDP) Model of Robot Team [8])** Given an indexed robot team $r_n \in R$ with an abstracted state set $S_n$, and each state $s \in S_n$ describes a local task performing state that robot team $r_n$ is exploring the target fort. The bounding overwatch task execution of robot team $r_n$ in a specific environment can be constructed as the MDP tuple

$$TE_n := (S_n, A_n, \delta_n, s_{0,n}, AP_n, \mathcal{L}_n, W_n),$$

where $A_n$ is an action set of the robot team; $\delta_n: S_n \times A_n \times S_n \to [0,1]$ describes the transition probability of robot team from state $s \in S_n$, to state $s' \in S_n$ by executing an action $\alpha \in A_n$ (i.e., the probability of robot team exploring from a destination fort to another by executing an action), and $\delta_n(s, \alpha, s') = Pr(s'|s, \alpha)$ ; $s_{0,n} \in S_n$ is the initial state; $AP_n$ is the set of atomic propositions of a set of task specifications; $\mathcal{L}_n: S_n \to 2^{AP_n}$ labels the robot states with the propositions derived from $AP_n$; and $W_n: S_n \times A_n \to R^+$ is the weight set.

In general, the process of the robot team exploring a specific fort is abstracted as a task performing state of the MDP. The MDP regulates the transition of an abstracted state with an action and labels the property of each abstracted state with predefined atomic propositions.

**Example 2.** Fig. 3. gives a graph representation of the MDP model of a robot team. States $s_1, s_2$ and $s_3$ are the task performing states representing the robot team is exploring $p_1, p_2$ and $p_3$ respectively. State $s_\epsilon$ is the idle state that the robot team is moving but not targeting for any fort, while state $s_f$ is the failure state that the robot team fails to move towards a desired fort with the corresponding action. The atomic propositions $p_1, p_2, p_3, p_f$ and $\varepsilon$ label the corresponding states in the figure.

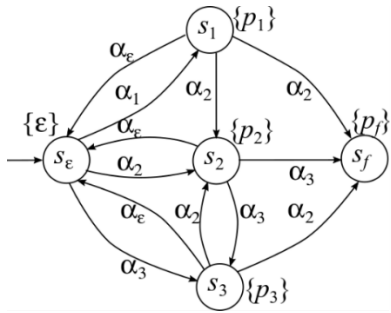

**Figure 3:** MDP model of the robot team in environment shown in Fig.1.

The defined MDP model of the robot team will need to be verified whether it can satisfy the bounding overwatch task specification. The verification results may be multiple task plans that can satisfy the task specification. Then, the most reliable task plan will need to be selected by referring to concrete metrics such as robot travelling distance, time and performance.

On the other hand, human-robot trust can be interpreted as the willingness of human to accept robot-produced information and robot's suggestions, thus to benefit from the advantages inherent in robotic systems, to assign tasks to robot, or to provide support to robot [21,22]. The impacting factors influencing human trust in robot can be categorized into the robot-related ones, such as robot performance and attribute, the environment-related ones, including feature of task environment and form of collaboration, and the human-related ones, such as human ability and characteristic.

For the bounding overwatch task considered in this paper, trust will mainly be used to evaluate the trustworthiness of the overwatch team members at a location in covering its remaining members that advance to the next bounding overwatch point. Therefore, a reliable bounding overwatch plan of the robot team to a key fort here will consider multi-dimensional metrics, such as line of sight and traversability, which affect the successful implementation of the task plan. We formulate the following problem.

**(Problem Setup)** Given multiple teams of heterogeneous robots for a bounding overwatch scenario. Let each robot team be associated with a capability MDP as defined in Def. 4 describing its bounding overwatch constraints in the environment. Develop a framework that can (1) guarantee a provably correct task plan for all the robot teams to satisfy a set of motion task specifications requested from human users; and (2) assign each robot team to satisfy the task plans with the highest trust level in performing the bounding overwatch task.

We propose the framework shown in Fig. 4 to solve the problem. More specifically, we will first use LTL, RE formulae or directly DFAs to describe the high-level global task specifications of multi-robot bounding overwatch in a terrain (see step 1). A parallel task decomposition process will first decompose the automaton (Def. 3) of the global task specification into parallel subtask automata (step 2). Thus, each parallel subtask automaton can be satisfied by a robot team independently and hence it improves the concurrency and efficiency of global task performing. Model checking techniques are used to synthesize the task plans satisfying each subtask automaton and capability MDP (Def. 4) of the robot team. The synthesized task plans are then combined with a discrete motion MDP (Def. 6) of each team to obtain the task and motion planning MDP (Def. 7) of the robot team (step 3). In the meantime, a dynamic computational trust model will be developed to evaluate robot trustworthiness in every bounding overwatch step. The trust of each robot team will be integrated with the task and motion planning MDP (step 4). Then, the most trustworthy bounding overwatch paths will be explored for each robot team to reach the desired key forts (step 5). Motion trajectories are then generated for each robot (step 6).

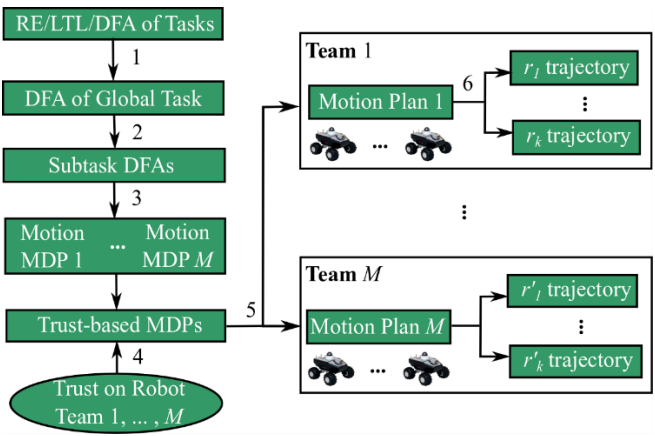

**Figure 4:** Flowchart of trust based SMP for MRS bounding overwatch.

## 3. TASK PLANNING FOR ROBOT TEAM BOUNDING OVERWATCH

In general, task requirements can be described by a variety of grammas [23]. In this paper, we require a human supervisor to provide a task specification in an LTL or RE format, which is relatively easier for human to formulate the task requirement and can be converted into the automaton [24]. More user-friendly grammars maybe developed for human supervisors so that they can describe and input their requirements [23].

The task specification in LTL or RE format is verified for the satisfaction of each robot team's capability in Def. 4. As the task specification mainly involves the reachability of task specification, we focus on the guarantee format of LTL for the task specification [18]. On the other side, we also aim to achieve the completion of local tasks in parallel processes with MRS, which improves the concurrency and efficiency of overall bounding overwatch.

### 3.1. Global Task Specification and Parallel Subtasks of Bounding Overwatch

User can assign multiple RE, LTL and DFA described bounding overwatch tasks for the robot teams. An overview of the temporal logic relation among and inside all these tasks can be achieved with a global task DFA, which can be synthesized with the regular language operations of the set of converted DFAs from each RE and LTL formula [17]. In this paper, we represent a synthesized global DFA of all the task specifications with $G_g = (X_g, E_g, f_g, x_{0,g}, X_{F,g})$ according to Def. 3. This global automaton describes all the bounding overwatch task plans satisfying human supervisor's input. Each robot team set is configured such that $2^{AP_n} \supseteq E_g$.

We aim to decompose the global bounding overwatch task specification $G_g$ into multiple parallel processes, which can improve the task performing concurrency. In [6], an iterated automaton decomposition algorithm is developed to decompose a global task DFA $G_g$ into a unique set of smallest subtask automata $G_1, \cdots, G_n, \cdots, G_N$, $n = 1, \cdots, N$. We adapt this algorithm here to the multi-robot bounding overwatch task for team allocation. As a result, we will have multiple parallel subtask automata $G_1, \cdots, G_n, \cdots, G_N$, which contain all the task plans to be performed in parallel processes. Then, the MRS can perform the assigned tasks in parallel processes instead of in a single process.

### 3.2. Validated Task Performing Plans of Bounding Overwatch

Assume each subtask automaton $G_n \coloneqq (X_n, E_n, f_n, x_{0,n}, X_{F,n})$. Each subtask automaton $G_n$ is assigned to a robot team $r_n$ with the capability MDP $TE_n \coloneqq (S_n, A_n, \delta_n, s_{0,n}, AP_n, \mathcal{L}_n, W_n)$ to satisfy the described local tasks. Here, the capability MDP model of the robot team satisfies the prerequisite that $2^{AP_n} \supseteq E_n$. Then, we can synthesize a product MDP to obtain the validated task plans for the complex high-level bounding overwatch task.

**Definition 5 (Bounding Overwatch Product MDP)** Given a subtask automaton $G_n \coloneqq (X_n, E_n, f_n, x_{0,n}, X_{F,n})$ describing the high-level bounding overwatch task specification, a robot team $R_n$ is modeled with an MDP $TE_n \coloneqq (S_n, A_n, \delta_n, s_{0,n}, AP_n, \mathcal{L}_n, W_n)$ describing the team's constraints in satisfying bounding overwatch task. A product MDP $TE_n \times G_n \coloneqq (S_n \times X_n, A_n, \delta'_n, (s_{0,n}, x_{1,n}), \mathcal{L}'_n, AP_n, W'_n)$ can be synthesized to present the validated results of task plans satisfying the task specification, where the state set $S_n \times X_n$ contains the robot and task completion states; the action set is $A_n$; the transition function $\delta'_n((s_n, x_n), a_n, (s'_n, x'_n))$ describes the transition probability from state $(s_n, x_n)$ to state $(s'_n, x'_n)$ with action $a_n$, $Pr((s'_n, x'_n)|(s_n, x_n), a_n) \neq 0$ if $\exists s_n \xrightarrow{a_n} s'_n$ and $x_n \xrightarrow{L(s'_n)} x'_n$; the initial state is $(s_{0,n}, x_{1,n})$ if $\exists x_{0,n} \xrightarrow{L(s_{0,n})} x_{1,n}$; the label relation $\mathcal{L}'_n: S_n \times X_n \to 2^{AP_n}$ satisfies $\mathcal{L}'_n(s_n, x_n) = L_n(s_n)$; and $W'_n: S_n \times X_n \times A_n \times S_n \times X_n \to R^+$ is the weight set.

We can obtain $N$ bounding overwatch product MDPs $TE_n \times G_n$ for a high-level task specification if provided with enough robot teams. Each product MDP presents all the validated task performing plans of bounding overwatch that can be implemented by the corresponding robot team independently.

## 4. TRUST ASSOCIATED MOTION PLANNING FOR ROBOT BOUNDING OVERWATCH

In Section 3, we obtained the bounding overwatch product MDPs that present all the validated task plans satisfying the task specification. Each task plan contains a sequence of bounding overwatch task performing states of a robot team. In this section, we aim to generate the most reliable task plan for the bounding overwatch of each robot team based on trust evaluation for each robot team's behavior. The successive bounding overwatch method is considered, i.e., the overwatch members will only need to reach the same region as that of the advance members in each bounding overwatch step. This method of bounding overwatch eases the control of cooperation between the team members. Then, the path planning of a robot team bounding overwatch can be regarded as equivalent to the planning for each single robot inside the team.

### 4.1. Motion Discretization of Robot Team Bounding Overwatch

We discretize the mission environment into cells after considering the constraints of robot size, robot sensing range as well as the terrain height. A single step of successive bounding overwatch process is conducted between a cell and one of its neighboring cells. This guarantees that the overwatch robot is always within the line of the sight of the bounding robot and hence the bounding robot can follow the overwatch robot. As a result, the exploration between two key forts can be decomposed into a sequence of bounding overwatch steps among the discretized cells. We abstract the motion process of each robot team $r_n$ among these cells as a discrete motion MDP $TM_n$.

**Definition 6. (Discrete Motion MDP)** Given a discretized environment $C_n$, construct a discrete motion MDP $TM_n$ for each robot team,
$$TM_n = (C_n, A_n^c, \delta_n^c, \mathcal{L}_n^c, S_n, R_n)$$
where the state set $C_n$ contains all the discretized cells that robot team can advance to; the action set $A_n^c$ contains all the actions that the robot team can take when deciding which neighboring cell it will advance to; the transition function $\delta_n^c(c_n, a_n^c, c_n') = Pr_n(c_n'|c_n, a_n^c)$ describes the transition probability from cell $c_n$ to $c_n'$ with action $a_n^c \in A_n^c$; $\mathcal{L}_n^c : C_n \to S_n$ labels whether each cell is the destination of the robot team's task performing state; the reward function $R_n : C_n \times A_n^c \to R^+$ quantifies the reliability of a transition with a value.

**Example 3.** A terrain can be discretized in the planner view as shown in Fig. 5. Each cell size sensing range of a robot. These cells are also labelled regarding whether they contain a destination fort. Reward function estimates the reliability of transition for a robot team from one cell to neighboring cell.

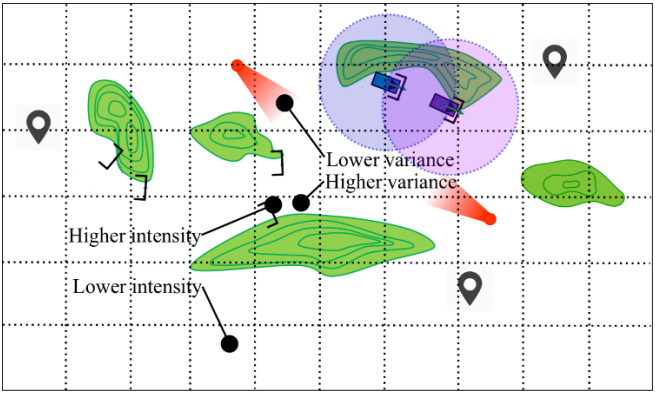

**Figure 5:** Discretized mission environment of Fig. 1. Each cell can be quantified by the arrangement of image intensities value to describe the traversability and by the image variance value to describe the line of sight.

The mission environment discretization enables an estimation on each small region of terrain regarding its trustworthiness for a robot team to conduct each bounding overwatch step.

Next, we compose the discrete motion MDP $TM_n$ with the product MDP $TE_n \times G_n$ to synthesize a task and motion planning MDP of robot team bounding overwatch.

**Definition 7. (Task and Motion Planning MDP)** Given the product MDP $TE_n \times G_n$ and discrete motion MDP $TM_n$, the composition of the two MDPs can be described as a task and motion planning MDP of a robot team bounding overwatch, which can be detailed as
$$PPM_n = \left(S_n^\psi, A_n^\psi, \delta_n^\psi, S_{0,n}^\psi, S_{F,n}^\psi, AP_n^\psi, \mathcal{L}_n^\psi, R_n^\psi\right)$$
where the state set is $S_n^\psi = S_n \times X_n \times C_n$; the action set is $A_n^\psi = A_n \times A_n^c$; the transition function $\delta_n^\psi\left(s_n^\psi, a_n^\psi, s_n^{'\psi}\right) = Pr\left((s_n', x_n')\middle|(s_n, x_n), a_n\right) \times Pr_n(c_n'|c_n, a_n^c)$ describes the transition probability from state $s_n^\psi = (s_n, x_n, c_n)$ to $s_n^{'\psi} = (s_n', x_n', c_n')$ with action $a_n^\psi = (a_n \times a_n^c)$; the initial state set is $S_{0,n}^\psi = \{(s_{0,n}, x_{1,n})\} \times C_n$; the accepted state set is $S_{F,n}^\psi = S_n \times X_{F,n} \times C_n$; the proposition set is $AP_n^\psi = AP_n' \times S_n$; the labeled relation is $\mathcal{L}_n^\psi : S_n^\psi \to AP_n^\psi$; and the reward function $R_n^\psi(s_n^\psi, a_n^\psi, s_n^{'\psi})$ returns a probabilistic value for a transition from state $s_n^\psi$ to state $s_n^{'\psi}$ with action $a_n^\psi$.

Given a state transition from time step $k$ to $k+1$, we can estimate the reward $R_n^\psi\left(s_n^{\psi,k}, a_n^{\psi,k}, s_n^{\psi,k+1}\right)$ with a predicted trust value at state $s_n^{\psi,k+1}$. The predicted trust value at state $s_n^{\psi,k+1}$ here can reflect the trustworthiness of a decision made by a robot team in the bounding overwatch task from state $s_n^{\psi,k} = (s, x, c)^k$ to $s_n^{\psi,k+1} = (s, x, c)^{k+1}$. The computation of the predicted trust value is introduced in the following subsection.

Proceedings of the 2020 Ground Vehicle Systems Engineering and Technology Symposium (GVSETS)### *4.2. Trust Evaluation of Robot Team Bounding Overwatch*

In this paper, we aim to obtain a trustworthy bounding overwatch process by considering the influence of terrain and exposure to potential adversaries on bounding overwatch. Therefore, we evaluate the traversability of robots in a terrain to estimate the influence of terrain on each robot team. The line of sight situation of each robot in the environment is utilized to estimate the capability of detecting risks of each team exposure to potential adversaries.

We use $g(\mathbf{y})$ to estimate the traversability cost of a given robot at a Cartesian position $\mathbf{y} = [y_1, y_2]^\top$ in the *x-y* plane. The traversability of the robot can be related to the terrain height, terrain texture, and mechanical limitations of the robot. In this paper, we assume homogeneous robots and associate the image intensities and image texture property value of each pixel in the height map with the corresponding position to evaluate traversability. Higher image intensity and image texture property value of a position indicates a tougher terrain for the motion of the robot team. We assume the traversability of a $\Delta \times \Delta$ pixels cell $c$ follows a normal distribution $g(c) \sim N(g_c, \zeta_{c,1})$, where $g_c$ is the mean value of the traversability costs of all the pixels in cell $c$, and $\zeta_c$ is the variance.

The line of sight of a cell can affect the sensing of the robot for surrounding environment and adversaries, which is related to the exposure risks of the robot team. The line of sight can be estimated with terrain height and sensing range of the robot. Here, we associate an image variance value of the sensing range of a robot at each position $\mathbf{y}$ to evaluate line of sight. Assume the image variance of the sensing range centered at a pixel follows a normal distribution $\sigma(c) \sim N(\sigma_c, \zeta_{c,2})$ in the $\Delta \times \Delta$ size cell of the height map. A higher mean value of the image variance of a cell corresponds to an overall complex terrain, which will make it difficult for a given robot to sense the surrounding situation. Thus, the image variance of the height map can be used to estimate the line of sight $\sigma(c)$ of the robot team at a cell $c$.

**Remark 1.** Consider the mission environment shown in Fig. 5. The image intensities, image texture property and variance values of each discrete cell can be estimated based on the 2D top view of the map. The cell with high value of line of sight is favorable for a robot to detect risks but may have disadvantages in traversability. Therefore, it needs to tradeoff between line of sight and traversability in selecting the bounding overwatch path. ●

Given a state $s_n^{\psi,k} = (s, x, c)^k$, denote the associated traversability $g_n^k = g(c^k)$ and line of sight $\sigma_n^k = \sigma(c^k)$. The associated trust value $\tau_n^k$ can be estimated based on the two costs $g_n^k, \sigma_n^k$ and the previous trust $\tau_n^{k-1}$. Denote $\mathbf{z}_n^k = [\tau_n^{k-1}, g_n^k, \sigma_n^k]^\top$. A more interpretable linear relation $\tau_n^k = \boldsymbol{\beta}^\top \mathbf{z}_n^k + \gamma^k$ is assumed to estimate the trust value $\tau_n^k$ with variables $\mathbf{z}_n^k$, where $\boldsymbol{\beta}^\top = [\beta_0, \beta_1, \beta_2]$, and $\gamma^k$ is the residue. Assume residual $\gamma^k \sim N(0, \xi^2)$, the trust value $\tau_n^k$ conditional on $\boldsymbol{\beta}, \mathbf{z}_n^k$ at a state $s_n^{\psi,k}$ satisfies

$$\tau_n^k | \boldsymbol{\beta}, \mathbf{z}_n^k, \xi^2 \sim N(\boldsymbol{\beta}^\top \mathbf{z}_n^k, \xi^2). \tag{1}$$

Then, we can predict the trust at a state $s_n^{\psi,k}$ with a dynamic Bayesian network (DBN), as shown in Fig 6.

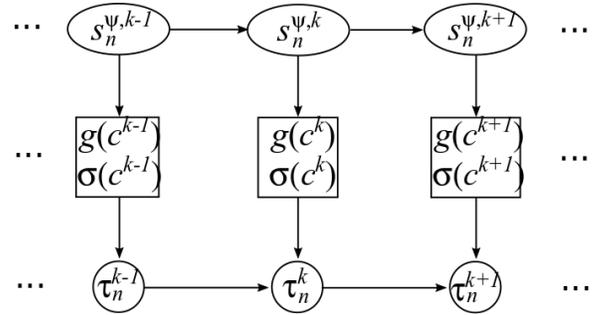

**Figure 6:** DBN based trust evaluation on task and motion planning.

Trust-based Symbolic Motion Planning for Multi-robot Bounding Overwatch, Huanfei Zheng, et al.

Page 8 of 13

$[z_n^0, z_n^1, \cdots, z_n^k]$. The full distribution of a trust value $\tau_n^t$ can be detailed as

$$Pr(\tau_n^k, \boldsymbol{\beta}, \mathbb{Z}_n^k, \xi^2) = Pr(\tau_n^k|\boldsymbol{\beta}, \mathbb{Z}_n^k, \xi^2)Pr(\boldsymbol{\beta}) \times Pr(g_n^k)Pr(\sigma_n^k)Pr(\tau_n^{k-1}, \boldsymbol{\beta}, \mathbb{Z}_n^{k-1}, \xi^2). \quad (2)$$

Thus, we can obtain $\tau_n^k \sim N(\bar{\tau}_n^k, \xi_n^k)$. Accordingly, we can obtain the expected trust value $\boldsymbol{E}(\tau_n^K) = \bar{\tau}_n^K$ of an arbitrary path $\rho_n = s_n^{\psi,0} \cdots s_n^{\psi,k} \cdots s_n^{\psi,K}$.

### 4.3. Optimal Path of MRS Bounding Overwatch

The trust evaluation on a path $\rho_n$ estimates the trustworthiness of a robot team in executing a task and motion plan satisfying the task specification. A task and motion plan with higher computational trust value has an overall higher traversability and line of sight considering the linear relation between the trust and the impacting factors. Therefore, we search for the most trustworthy task and motion planning path from the synthesized MDP in Def 7. The trustworthiness of every path of the MDP will need to be estimated so that an optimal one can be finally obtained. We use the Dijkstra search strategy to explore the transitions of the task and motion planning MDP $PPM_n$. At each step, we search for the state that has the maximum expected trust value, which can be evaluated with Eqn. (2). The algorithm is shown in Alg. 1.

Algorithm 1. Optimal Motion Plan of MRS

**Function** $OptmPath(PPM_n)$

1: Set $Q_n = S_n^\psi$
2: Initialize $V(s_n^\psi) = 0, \forall s_n^\psi \in Q_n$
3: Initialize $\tau_n^0$
4: Value function $V(s_{0,n}^\psi) = -\boldsymbol{E}(\tau_n^0)$
5: **While** $Q_n$ is not empty:
6:     Obtain $s_n^\psi \leftarrow argmin_{s_{u,n}^\psi \in Q_n} V(s_{u,n}^\psi)$
7:     Remove $s_n^\psi$ from $Q_n$
8:     **for** $s_{v,n}^\psi \in SuccessorOf(s_n^\psi)$:
9:         Obtain $z_n^k$ related to $s_n^{\psi,k} = s_n^\psi$
10:         $\tau_{v,n} \leftarrow \boldsymbol{\beta}^\top z_n^k + \gamma^k$
11:         **if** $-\boldsymbol{E}(\tau_{v,n}) < V(s_{v,n}^\psi)$:
12:             Assign $V(s_{v,n}^\psi) = -\boldsymbol{E}(\tau_{v,n})$
13:             $PrecessorOf(s_{v,n}^\psi) = s_n^\psi$
14:         **end if**
15:     **end for**
16: **end While**
17: Obtain $s_{F,n}^\psi \leftarrow argmin_{s_{F,n}^\psi \in Q_n} V(s_{F,n}^\psi)$
18: Initialize $\rho_n = [\,], s_n^\psi = s_{F,n}^\psi$
19: **While** $\exists PrecessorOf(s_n^\psi) \vee s_n^\psi = s_{0,n}^\psi$:
20:     Update $\rho_n = [\rho_n\ s_n^\psi]$
21:     Update $s_n^\psi = PrecessorOf(s_n^\psi)$
22: **end While**
23: **return** $\rho_n$

The most trustworthy path is a sequence of discrete cells $c$ for the robot team to advance to. Trajectory of each robot can be synthesized based on the sequence of discrete cells given an omnidirectional mobile robot. The robots always seek to reach the centroid of the next adjacent cell in the discrete path. This guarantees the robots will always explore the planned region while moving between adjacent regions in the planned path.

## 5. ROS GAZEBO SIMULATION OF MRS BOUNDING OVERWATCH

We simulate the trust based SMP framework for MRS bounding overwatch in a terrain shown in Fig. 7, where $F_1 - F_6$ are the six forts of interest for the bounding overwatch task. A complicated task specification needs to be satisfied: (1) "advance to fort $F_3$ first, then conquer $F_4$ or $F_2$"; (2) "advance to $F_4$ first, then conquer $F_3$"; (3) "conquer $F_6$ after either requirement (1) or (2) is completed".

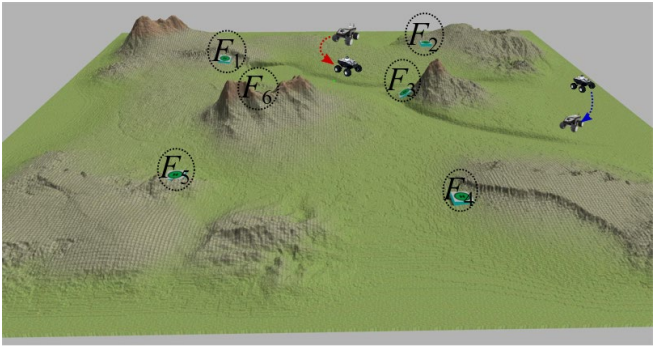

**Figure 7:** Perspective view of the bounding overwatch terrain

Task requirements (1) and (2) can be easily described with RE $f_3(f_4 + f_2)$ and $f_4 f_3$, respectively. They can also be converted to their equivalent DFAs. The global view of the three task requirements can be described with the automaton operation on these DFA, shown in Fig. 8. We also add a backward transition at state 3 considering that it may need to advance back to $F_3$ or $F_4$ in case of any accidents.

The decomposability of global task DFA $G_g$ is verified according to the algorithm introduced in Sec. 3.1. The decomposition results are subtask automata $G_1$ and $G_2$ shown in Fig. 8 (right). They are two parallel subtask automata that can be satisfied in two independent processes.

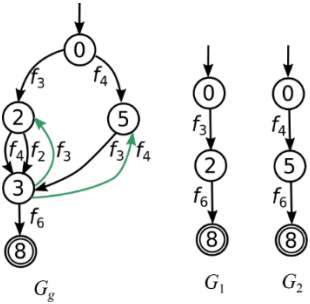

**Figure 8:** Automaton of the task requirements

Therefore, we configure two robot teams, each consisted of two Summit-XL robots shown in Fig. 7, to complete the bounding overwatch task in parallel. As shown in Fig. 7, we label the robot teams as red and blue, respectively. The two robot teams are given by the following capability MDPs in Fig. 9, which describe their constraints of task capability in the terrain according to Def. 4. The MDP $TE_1$ is for red robot team, while $TE_2$ is for blue robot team. Each state $s_i$ of $s_1 - s_6$ represents the corresponding robot team is exploring for the destinated fort $F_i$. State $s_\varepsilon$ represents the roaming state without targeting any fort, while state $s_f$ denotes the error state that robot team fails to work. Atomic propositions $f_i$, $f_{err}$ and $\varepsilon$ label each state so that the MDPs can be related with the task specifications.

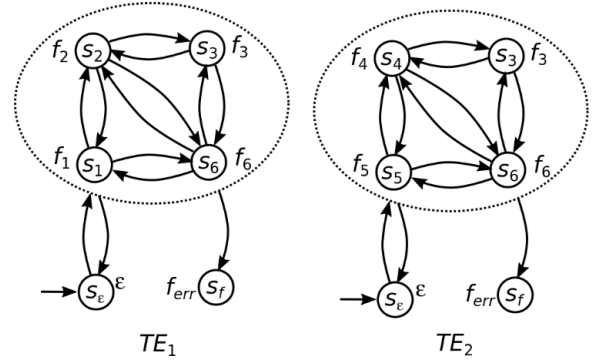

**Figure 9:** The capability MDP models of the robot teams for the bounding overwatch in the terrain. Note the directed edges between states $s_\varepsilon$, $s_f$ and dashed circle describe that all the states in each circle can transit to $s_\varepsilon$, $s_f$ with the shown transition relations.

A provably correct task plan in the current environment can be obtained from the bounding overwatch product MDP that is synthesized with subtask automaton $G_n$ and MDP $TE_n$ of robot team according to Def. 5, where $n = 1,2$. The transition description of the two product MDPs $TE_n \times G_n$ are shown in Fig 10. They give the results regarding the task plans that can satisfy both the task specification and robot capability. $TE_1 \times G_1$ shows the correct task planning results that the red robot team can (1) explore fort $F_6$ after the team completes exploring fort $F_3$, or (2) first enter a roaming state $s_\varepsilon$ and then explore fort $F_6$, after the team completes exploring fort $F_3$. The similar explanation works for blue team with $TE_2 \times G_2$, i.e., either (1) visiting fort $F_6$ after visiting fort $F_4$ or (2) entering roaming state and then visiting fort $F_6$, after visiting fort $F_4$.

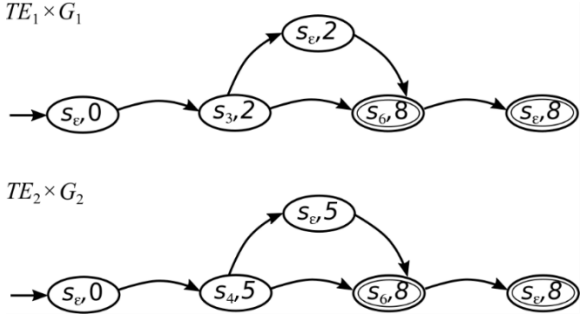

**Figure 10:** The product MDPs of the robot teams for the bounding overwatch in the terrain.

The product MDPs only present the correct task plans for robot team to satisfy the task specification. Next, we discretize the environment and develop a motion MDP for each robot according to Def. 6. The graphic view is shown as the discrete cells in Fig. 11. We compose the robot motion MDP with the product MDP. The traversability and line of sight of each discrete cell can be estimated based on the discrete environment. This enables the estimation on trust value of each state. Here, we seek to obtain a discrete path with highest mean trust value. We then use the Dijkstra algorithm in Alg. 1 to find the discrete task and motion plan with the maximum mean trust value among all the plans. We also generate the paths that consider best traversability, or best line of sight, respectively in Fig. 11 (a), (b). Here, Fig 11 (b) presents the map of line of sight value instead of the original map, and the dashed rectangle areas are the sensing range of each team. In addition, each cell with over low traversability is labelled with a cross.

We simulated with different weights $\boldsymbol{\beta} = [\beta_0, \beta_1, \beta_2]^\mathsf{T}$ of the computational trust model to generate the paths of bounding overwatch. Fig. 11 (c) are the generated representative paths with weights $\boldsymbol{\beta} \sim N(\boldsymbol{\mu}, \boldsymbol{\zeta})$, $\boldsymbol{\mu} = [0.27, 0.33, 0.40]^\mathsf{T}$, $\boldsymbol{\zeta} = [[0.01, -0.01, -0.01]; [-0.01, 0.01, 0]; [-0.01, 0, 0.01]]$. The probabilistic trust values of the two robot teams for their task and motion plans are evaluated at each step of the discrete path, as shown in Fig. 12.

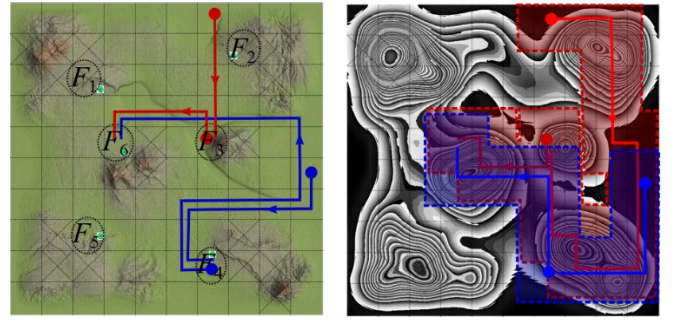

(a) Paths providing the best traversability for the two robot teams.

(b) Paths providing the best line of sight for the two robot teams.

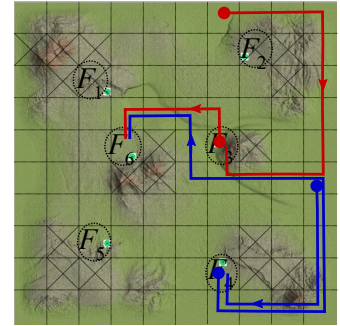

(c) The most trustworthy paths for the two robot teams.

**Figure 11:** Task and motion plans of the two robot teams for the bounding overwatch task in the terrain.

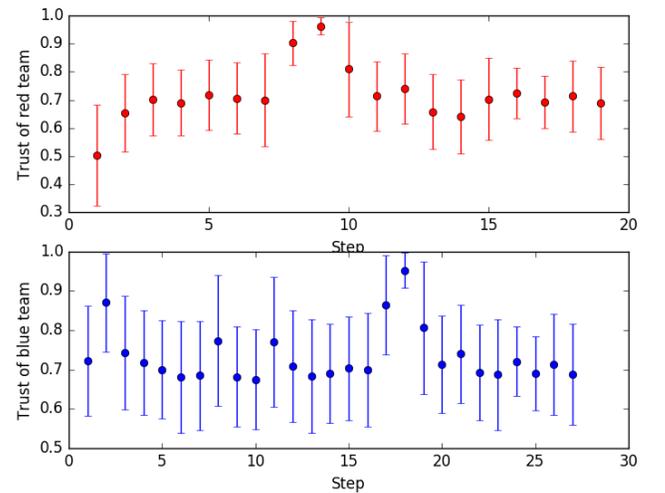

**Figure 12:** Estimated trust values of the robot teams for the most trustworthy task and motion plans

## 6. CONCLUSION

We developed a trust based SMP framework for MRS to satisfy complex bounding overwatch tasks constrained by temporal logics. It aims to satisfy a bounding overwatch task specification in safe and reliable approach. Multiple task and motion planning MDPs were first generated to guarantee both correctness and parallel execution of the complex bounding overwatch tasks by the MRS. A computational trust model was then constructed with the traversability and line of sight of robots in the mission terrain. The trust evaluation was integrated into the task and motion MDP and utilized to predict the trustworthiness of each robot team's potential behavior in executing a task plan. The most trustworthy task and motion plan is explored with a customized Dijkstra searching algorithm. The task and motion plan presented a reliable MRS bounding overwatch process which trades off the traversability and safety of robot motion in a terrain.